\documentclass{article}
\usepackage{spconf,amsmath,graphicx}
\usepackage{url}

\title{High Fidelity Fingerprint Generation: Quality, Uniqueness, and Privacy}
\twoauthors{Keivan Bahmani, Richard Plesh, Peter Johnson, Stephanie Schuckers \thanks{This material is based upon work supported by the Center for Identification Technology Research and the National Science Foundation under Grant No.$650503$.}}
	{Clarkson University\\
	Potsdam, NY, USA}
  {\hspace{-0.6in}Timothy Swyka}
	{\hspace{-0.6in}Precise Biometrics\\
    \hspace{-0.6in}Potsdam, NY, USA}

\usepackage{fancyhdr}

\begin{document}
\maketitle
\begin{abstract}
In this work, we utilize progressive growth-based Generative Adversarial Networks (GANs) to develop the Clarkson Fingerprint Generator (CFG). We demonstrate that the CFG is capable of generating realistic, high fidelity, $512\times512$ pixels, full, plain impression fingerprints. Our results suggest that the fingerprints generated by the CFG are unique, diverse, and resemble the training dataset in terms of minutiae configuration and quality, while not revealing the underlying identities of the training data. We make the pre-trained CFG model and the synthetically generated dataset publicly available at \url{https://github.com/keivanB/Clarkson_Finger_Gen} %
\end{abstract}

\section{Introduction}
\label{sec:intro}
 Fingerprint identification systems are widely accepted as inexpensive and secure, making them one of the most prevalent forms of biometric recognition. Current state-of-the-art fingerprint recognition systems heavily rely on Convolutional Neural Networks (CNNs) \cite{sundararajan_deep_2018}. While these systems show exceptional performance, they require expensive large-scale fingerprint datasets for training and evaluation. However, collecting and sharing large-scale biometric datasets comes with inherent risks and privacy concerns. For instance, the National Institute of Standards and Technology (NIST) recently discontinued several publicly available datasets from their catalog due to privacy issues \cite{website}. Generating synthetic fingerprints can help alleviate both expense and privacy issues. Synthetic datasets can be generated easily at scale while still representing the training dataset and shielding the identity of the individuals that were used during training.

The traditional approach to fingerprint synthesis involves sampling from independent statistical models for orientation field and minutiae with Gabor-filtering or other models to generate the final ridge structure \cite{cappelli_sfinge_2004, johnson_texture_2013, zhao_fingerprint_2012}. The fingerprints generated using these approaches suffer from some shortcomings. The additive noise used in the ridge valley generation process can give the generated fingerprints a distinct visual pattern. Additionally, fingerprints generated using the traditional approaches can be distinguished from real fingerprints using their minutiae distribution \cite{gottschlich_separating_2014}. Finally, the independent modeling used in the traditional approach is not necessarily able to capture the correlation between the orientation field, ridge valley structure, and minutiae patterns.

\begin{figure}[h]
\begin{minipage}[b]{0.2\linewidth}
  \centering
  \centerline{\includegraphics[width=2.5cm]{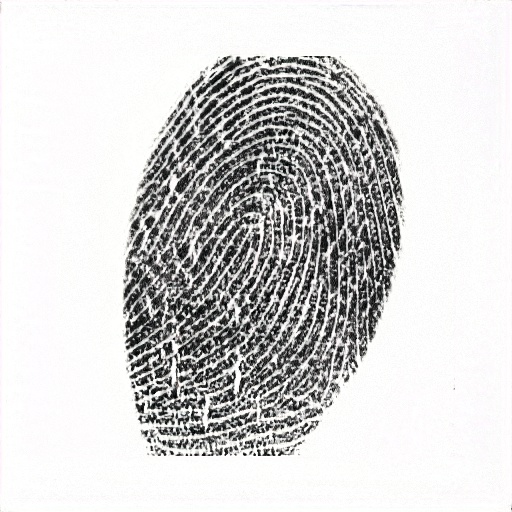}}
\end{minipage}
\hfill
\begin{minipage}[b]{0.2\linewidth}
  \centering
  \centerline{\includegraphics[width=2.5cm]{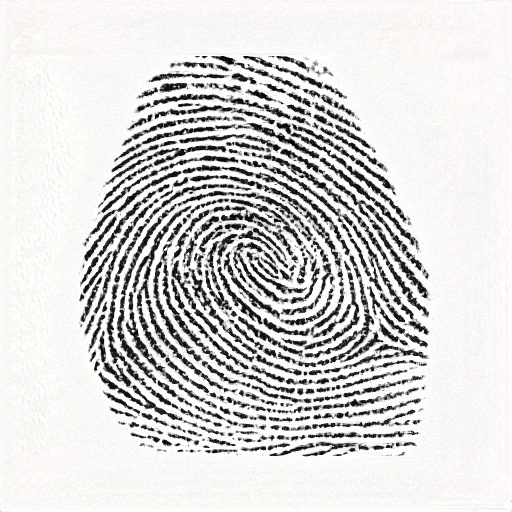}}
\end{minipage}
\hfill
\begin{minipage}[b]{0.2\linewidth}
  \centering
  \centerline{\includegraphics[width=2.5cm]{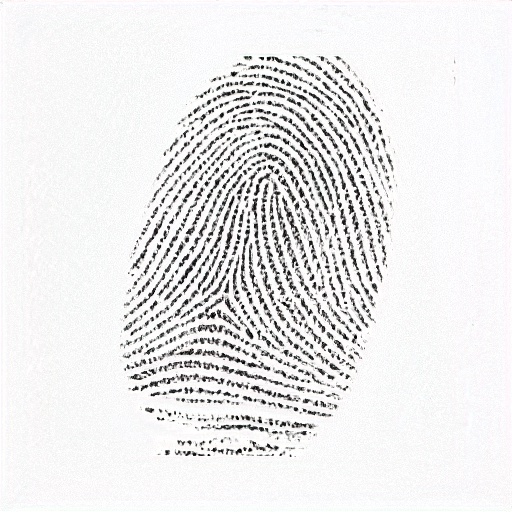}}
\end{minipage}
\hfill
\begin{minipage}[b]{0.2\linewidth}
  \centering
  \centerline{\includegraphics[width=2.5cm]{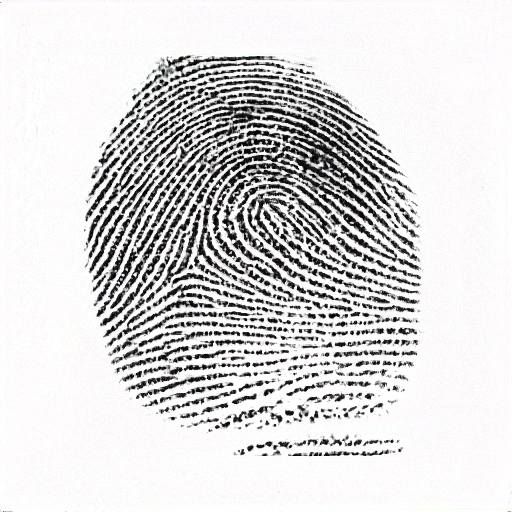}}
\end{minipage}
\begin{minipage}[b]{0.2\linewidth}
  \centering
  \centerline{\includegraphics[width=2.5cm]{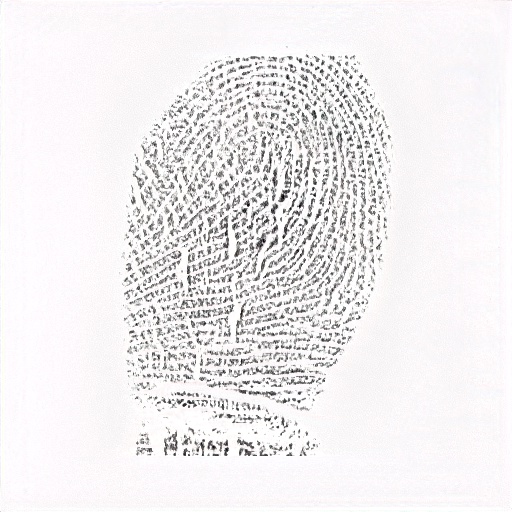}}
\end{minipage}
\hfill
\begin{minipage}[b]{0.2\linewidth}
  \centering
  \centerline{\includegraphics[width=2.5cm]{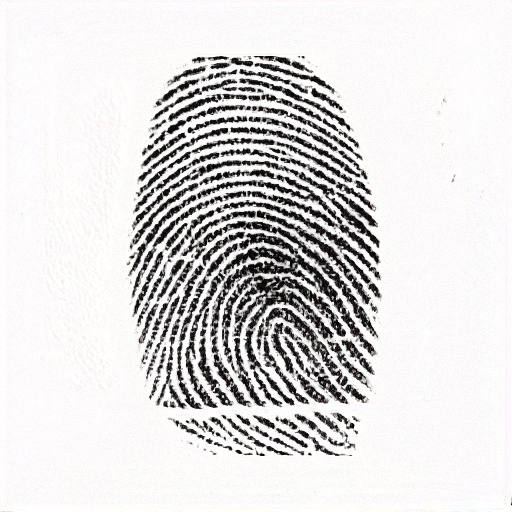}}
\end{minipage}
\hfill
\begin{minipage}[b]{0.2\linewidth}
  \centering
  \centerline{\includegraphics[width=2.5cm]{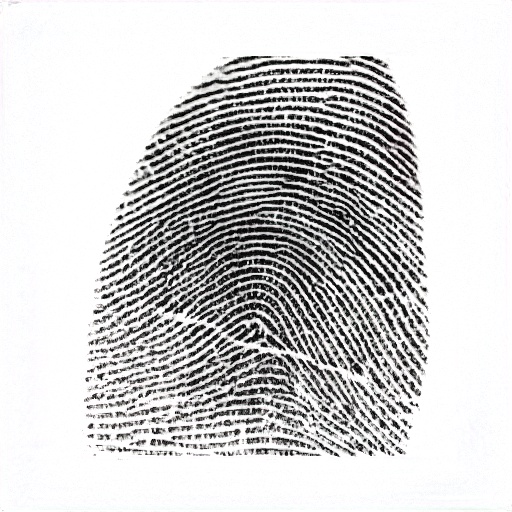}}
\end{minipage}
\hfill
\begin{minipage}[b]{0.2\linewidth}
  \centering
  \centerline{\includegraphics[width=2.5cm]{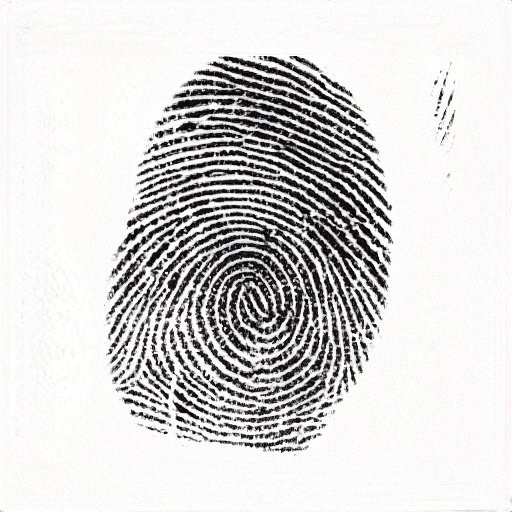}}
\end{minipage}
\caption{Examples of generated fingerprint images by the proposed approach ($512\times512$ pixels at 500 dpi)}
\label{fig:samples}
\end{figure}

More recently, GAN models have been employed for fingerprint synthesis \cite{bontrager_deepmasterprint:_2017, cao_fingerprint_2018, minaee_finger-gan_2018, fahim_lightweight_2020}. Contrary to the previous fingerprint synthetic approaches, GANs don't rely on independent statistical models for each aspect of fingerprints. They are capable of learning the high-dimensional probability distribution of the training data and generating samples from the learned distribution. Previous GAN-based fingerprint synthesis work mainly utilized the Improved Wasserstein GAN (IWGAN) architecture \cite{gulrajani_improved_2017-1}. These models, however, are unstable and fail to generalize to high-resolution images \cite{cao_fingerprint_2018, minaee_finger-gan_2018, fahim_lightweight_2020}. Previous work has proposed different training regimens to address these issues. Finger-GAN utilizes additional total variation constraints to impose connectivity within the generated images \cite{minaee_finger-gan_2018}. Fahim et al. proposed a loss doping approach to stabilize the training process and prevent mode collapse \cite{fahim_lightweight_2020}. The proposed methods improved the stability of the training process, but failed to produce full plain impression fingerprints with precise boundary and high fidelity. Cao et al. focused on fingerprint search at scale and utilized a large-scale dataset of 250,000 rolled fingerprints. Despite the large dataset, the authors still observed the same mode collapse issues. In response, they proposed an unsupervised pre-training step, adding additional computational overhead. To the best of our knowledge, this is the only GAN-based model in previous work that is capable of producing high fidelity, $512\times512$ pixels, 500 dpi, rolled fingerprints \cite{cao_fingerprint_2018}. In this work, we introduce the Clarkson Fingerprint Generator (CFG), a GAN-based fingerprint synthesis model which uses progressive growth training to generate realistic $512\times512$ pixels, plain impression fingerprint images, Figure \ref{fig:samples} illustrates examples of the synthetic fingerprints. Our contributions are as follows:

\begin{itemize}

\item Contrary to the previous IWGAN-based fingerprint synthesis models, the CFG utilizes a multi-resolution and progressive growth training approach \cite{karras_progressive_2017, karras_style-based_2019}. The CFG can generate high fidelity plain fingerprint with realistic shape and boundaries at $512\times512$ pixels and does not suffer the mode collapse and quality issues associated with previously proposed IWGAN-based fingerprint generators \cite{minaee_finger-gan_2018, fahim_lightweight_2020}. When compared to the model used in \cite{cao_fingerprint_2018}, the multi-resolution fingerprint synthesis model can be trained using a smaller dataset and without computationally expensive pre-training steps.

\item We compare quality metric distributions to assess diversity of the synthetic fingerprints and their similarity to bonafide fingerprints. We also match every synthetic fingerprint to every bonafide fingerprint to ensure that the synthetic fingerprints do not reveal the real identities.

\item We utilized a CNN-based Presentation Attack Detection (PAD) model to evaluate fingerprints generated using the CFG. This process reaffirms the high fidelity of the samples generated using the CFG.

\item We make the pre-trained CFG model and the synthetically generated fingerprints publicly available. To the best of our knowledge, the CFG is the first publicly available GAN-based fingerprint synthesis model.
\end{itemize}

\section{Clarkson Fingerprint Generator}
\label{sec:Proposed}
In this work, we utilize multi-resolution training for fingerprint synthesis \cite{karras_style-based_2019}. Multi-resolution GAN models start the training process by training both the Generator (G) and Discriminator (D) at lower spatial resolutions and progressively increasing (growing) the spatial resolution throughout the training. Progressive growth-based GANs are capable of effectively capturing high-frequency components of the training data and producing high-fidelity and realistic human faces \cite{karras_style-based_2019}. The main known limitation of the progressive growth approach is the generator's strong location preference for details. This issue has led to artifacts in generating high-resolution faces across different poses \cite{karras_analyzing_2020}. However, fingerprint recognition systems operate at a relatively fixed scale and do not suffer from the pose, illumination, and expression variations associated with face images. As a result, we believe this architecture alleviates the problems observed in the previous IWGAN-based fingerprint synthesis models while introducing a minimal amount of artifacts to the synthesized fingerprints. To the best of our knowledge, the CFG is the first fingerprint synthesis model that leverages multi-resolution and progressive growth training.

\section{Training and Evaluation}
\label{sec:Training}
The CFG utilizes StyleGAN architecture \cite{karras_style-based_2019}. The model is trained from scratch using 72,000, $512\times512$ pixels bonafide fingerprints from 250 unique identities, captured using a Crossmatch Guardian scanner (DB-1). The CFG is trained in an unsupervised manner, i.e. we did not provide the model with unique identity labels. The bonafide fingerprints are processed at 8 resolutions from $4\times4$ up to $512\times512$ pixels during progressive training. Subsequently, we utilized the CFG to generate 50,000 synthetically generated $512\times512$ plain impression fingerprints (DB-2). Note that the synthetic fingerprints are generated without any truncation to represent the full range of the CFG \cite{karras_style-based_2019}. 

Previous work on fingerprint synthesis relied on Frechet Inception Distance (FID) \cite{minaee_finger-gan_2018} and Structural SIMilarity (SSIM) \cite{fahim_lightweight_2020} measures to respectively evaluate the quality and diversity of the generated samples. In the test, the CFG achieved FID of 24 which is considerably better than the FID of 70 reported by Finger-GAN \cite{minaee_finger-gan_2018}. We calculated FID using the 50,000 fingerprints. Unfortunately, the use of in-house datasets and proprietary code limits our ability to compare FID with previous fingerprint synthesis models. Also, since FID utilizes an inception network trained on ImageNet \cite{huang_global_2006}, it is best suited for evaluating generators of natural images rather then biometric images. Consequently, we utilize the BOZORTH3 minutiae-based fingerprint matcher \cite{ko_users_2007} to evaluate the uniqueness of the synthetically generated fingerprints through their imposter distribution \cite{cao_fingerprint_2018} and expand upon previous works by evaluating the quality and diversity of the synthetic fingerprints through fingerprint metrics. We evaluated the quality of the fingerprints using NIST NFIQ 2.0 \cite{tabassi_nfiq_2016} and utilized the NIST NBIS software \cite{ko_users_2007} to evaluate and compare the minutiae configuration of the training (DB-1) and synthetic (DB-2) fingerprints. Additionally, we leveraged the work of Olsen et al. to extract features based on ridge-valley signature \cite{olsen_finger_2016}. To accurately estimate the ridge-valley features, each fingerprint is decomposed into overlapping blocks of $32\times32$ pixels and we averaged the results over the 15 patches with lowest standard deviation in terms of ridge valley uniformity (highest quality).

Additionally we evaluate the synthetic fingerprints through the use of a PAD model trained on a separate set of 50,000 (representing 250 people) and 35,000 PA images (representing 11 PA types) using Mobile NASNet architecture \cite{zoph_learning_2018}. We replace the fully connected layers with four layers, three RELU activated layers of size 500, 50, 10, and a 2-unit softmax layer representing bonafide vs PA fingerprints. The PAD model is initialized using weights trained on ImageNet and subsequently trained using full fingerprint images. We utilized the Adam optimizer \cite{kingma_adam:_2014}, a learning rate of 0.00001, and batch size of 128. The NASNet model achieves $13.5\%$ Attack Presentation Classification Error Rate (APCER) at $0.5\%$ Bonafide Presentation Classification Error Rate (BPCER).

The training of CFG, the CNN-based PAD model, and fingerprint synthesis process is carried out using 4 NVIDIA P100 GPUs provided by the National Science Foundation (NSF) Chameleon testbed \cite{keahey_lessons_2020}.

\section{RESULTS AND DISCUSSIONS}
\label{sec:Results}
Figure \ref{fig:samples} depicts several synthetic fingerprints generated with the CFG. Upon visual investigation, the CFG generates diverse, high-fidelity full plain impression fingerprints at $512\times512$ pixels without noisy boundaries and artifacts observed in the IWGAN-based fingerprint synthesis models \cite{fahim_lightweight_2020, minaee_finger-gan_2018}. 

\begin{table*}[ht!]
\centering
\begin{tabular}{|c| c| c| c| c| c| c| c| c|} 
\hline
& \multicolumn{4}{|c|}{DB-1 (Bonafide)} & \multicolumn{4}{|c|}{DB-2 (Synthetic)} \\
\hline
Measure & Mean & STD & Skewness & Kurtosis &  Mean & STD & Skewness & Kurtosis \\
\hline\hline
Ridge Ending Minutiae Count \cite{ko_users_2007} & 47.378 & 15.019 & 1.040 & 2.323 & 45.606 & 12.371 & 0.662& 0.602\\
\hline
Bifurcation Minutiae Count \cite{ko_users_2007}& 19.723 & 11.831 & 1.706 & 5.787 & 21.11 & 11.247 & 1.087 & 0.994\\
\hline
Reliability of Ridge Minutiaes \cite{ko_users_2007}& 0.458 & 0.098 & 0.149 & -0.048 & 0.453 & 0.081 & -0.148& -0.326\\
\hline
Reliability of Bifurcation Minutiaes \cite{ko_users_2007}&  0.592& 0.142 & -0.547 & 0.005 & 0.626 & 0.145 & -0.668& -0.032\\
\hline
Percentage of Bifurcation Minutiaes \cite{ko_users_2007}& 0.285 & 0.120 & 0.518 & 0.192 & 0.304 & 0.097 & 0.267& -0.043\\
\hline
Ridge Count \cite{olsen_finger_2016} & 3.272 & 0.375 & 0.420 & 0.66 & 3.331 & 0.334 & 0.326& -0.36\\ 
\hline
White Lines Count \cite{olsen_finger_2016} & 3.271&  0.376& 0.423 & 0.623 & 3.323 & 0.337& 0.353& -0.356\\
\hline
RTVTR  \cite{olsen_finger_2016} & 0.921&  0.205 & 0.657 & 0.446 & 0.910 & 0.148 & 0.387& -0.205\\
\hline
Area of the Fingerprint & 95.392& 24.224 & 0.349 & -0.317 & 94.019 & 21.232 & 0.688& 0.307\\
\hline
NFIQ2 Score  \cite{tabassi_nfiq_2016} & 57.406 & 18.482  & -0.365 & 2.622 & 54.687 & 14.616& -0.499 & 1.058\\
\hline
\end{tabular}
\caption{Fingerprint metrics for bonafide (DB-1) and synthetic (DB-2) fingerprints. NFIQ2 score has a range of $[0,100]$ and The Ridge to Valley Thickness Ratio (RTVTR) has a range of $[0,1]$.}  
\label{table:Finger_Metrics}
\end{table*}

\subsection{Matching Performance}
\begin{figure}[ht]
\begin{minipage}[b]{.48\linewidth}
  \centering
  \centerline{\includegraphics[width=4.3cm]{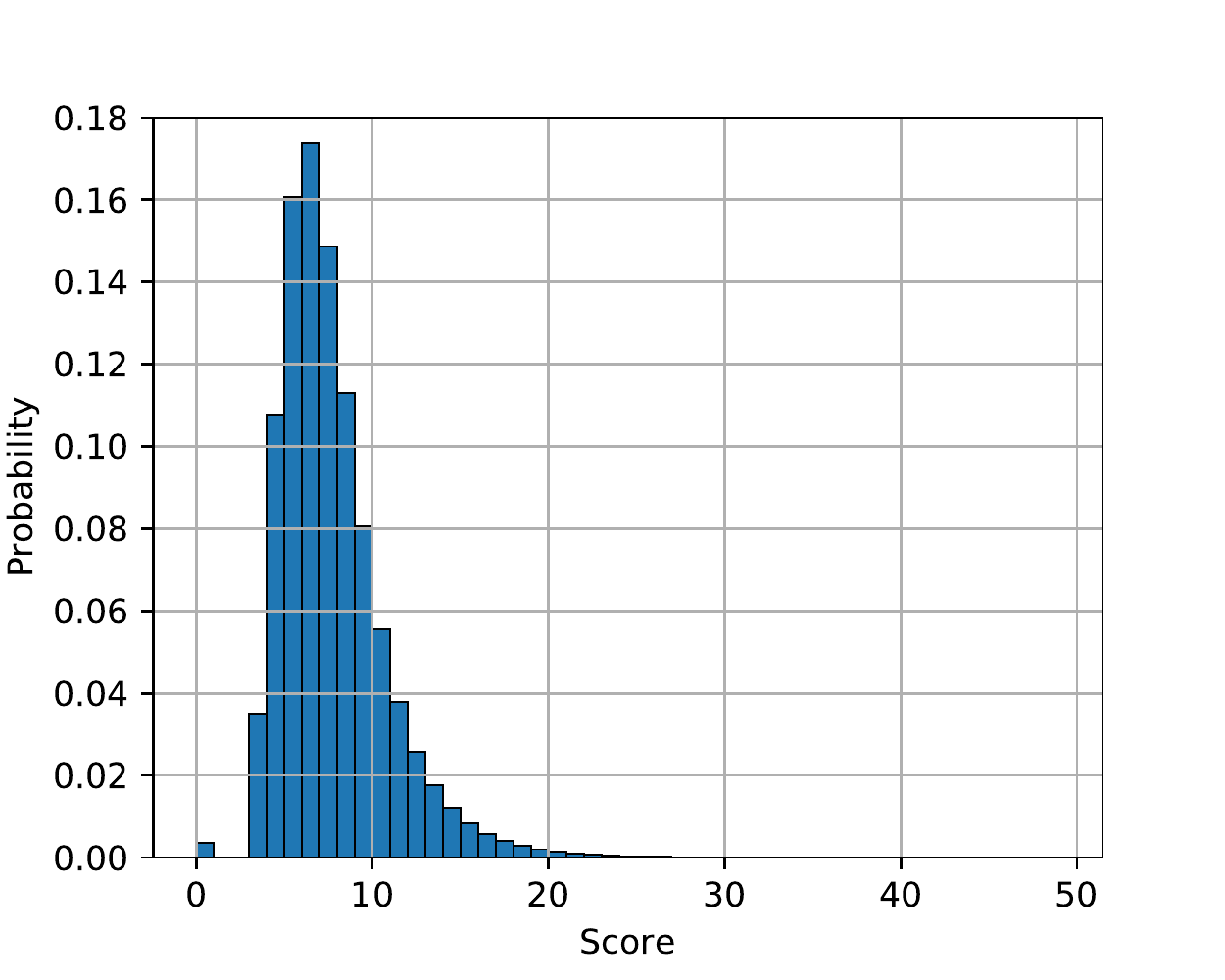}}
  \centerline{(a) Bonafide}\medskip
\end{minipage}
\hfill
\begin{minipage}[b]{0.48\linewidth}
  \centering
  \centerline{\includegraphics[width=4.3cm]{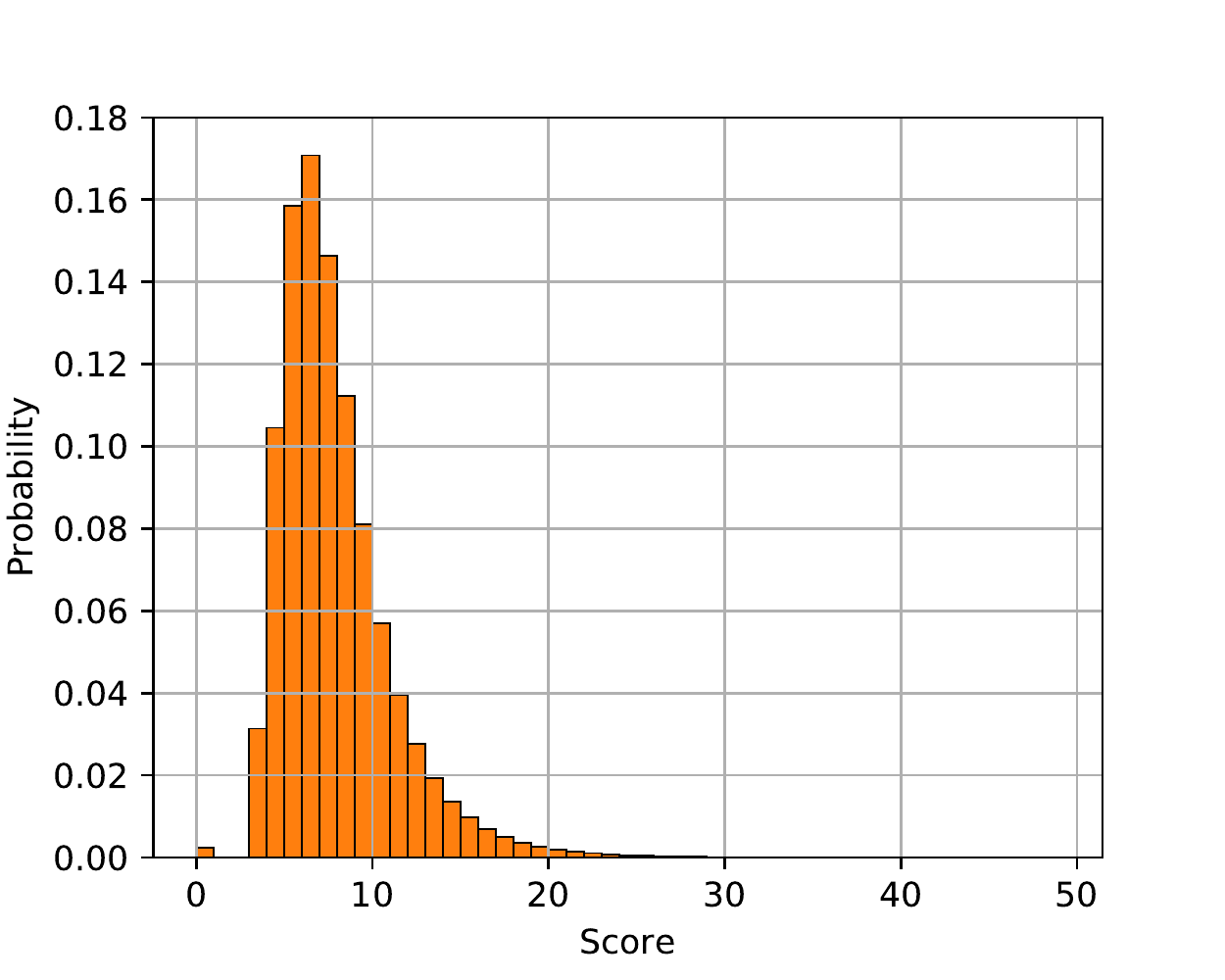}}
  \centerline{(b) Synthetic v. Bonafide }\medskip
\end{minipage}
\begin{minipage}[b]{.48\linewidth}
  \centering
  \centerline{\includegraphics[width=4.3cm]{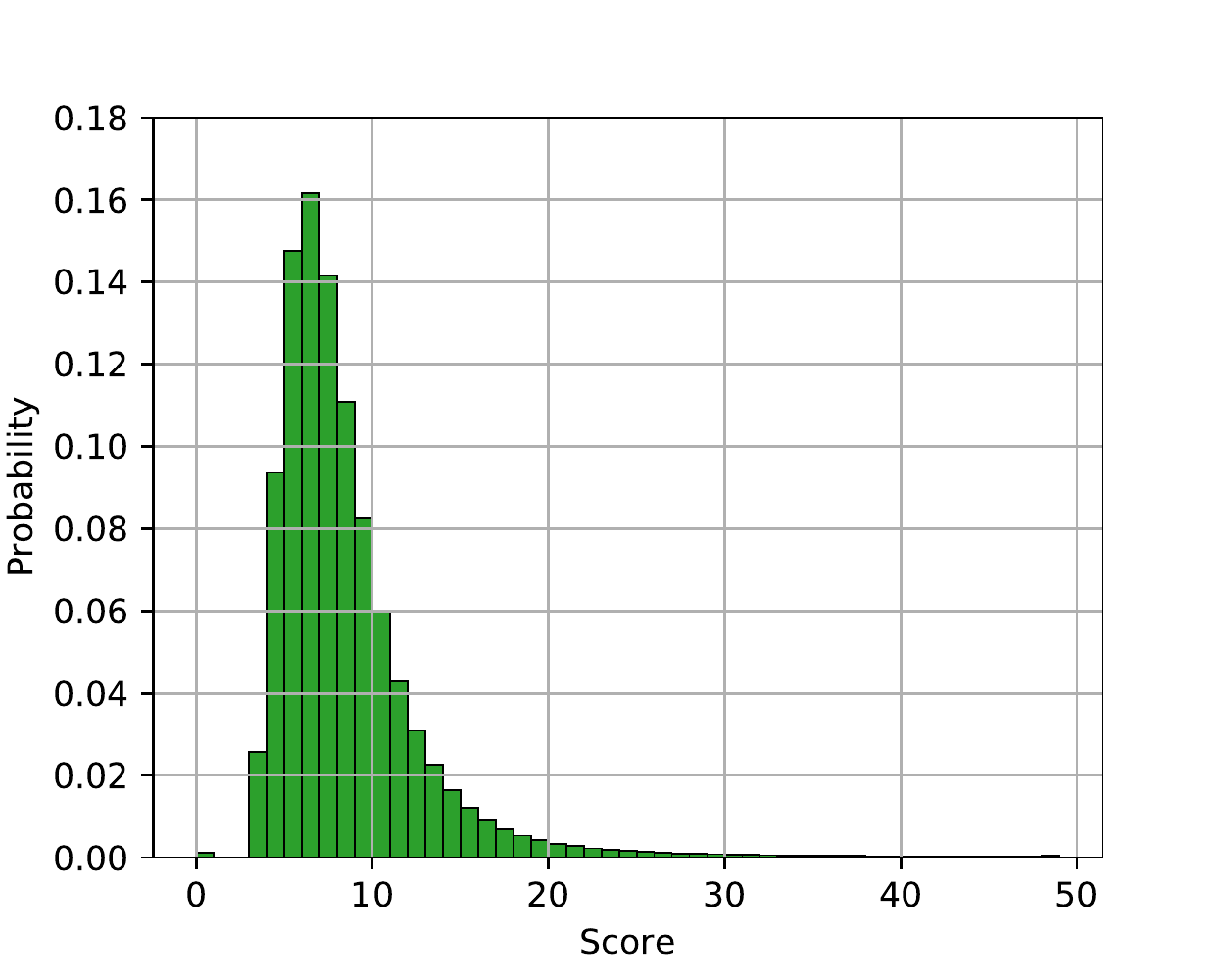}}
  \centerline{(c) Synthetic }\medskip
\end{minipage}
\hfill
\begin{minipage}[b]{0.48\linewidth}
  \centering
  \centerline{\includegraphics[width=4.3cm]{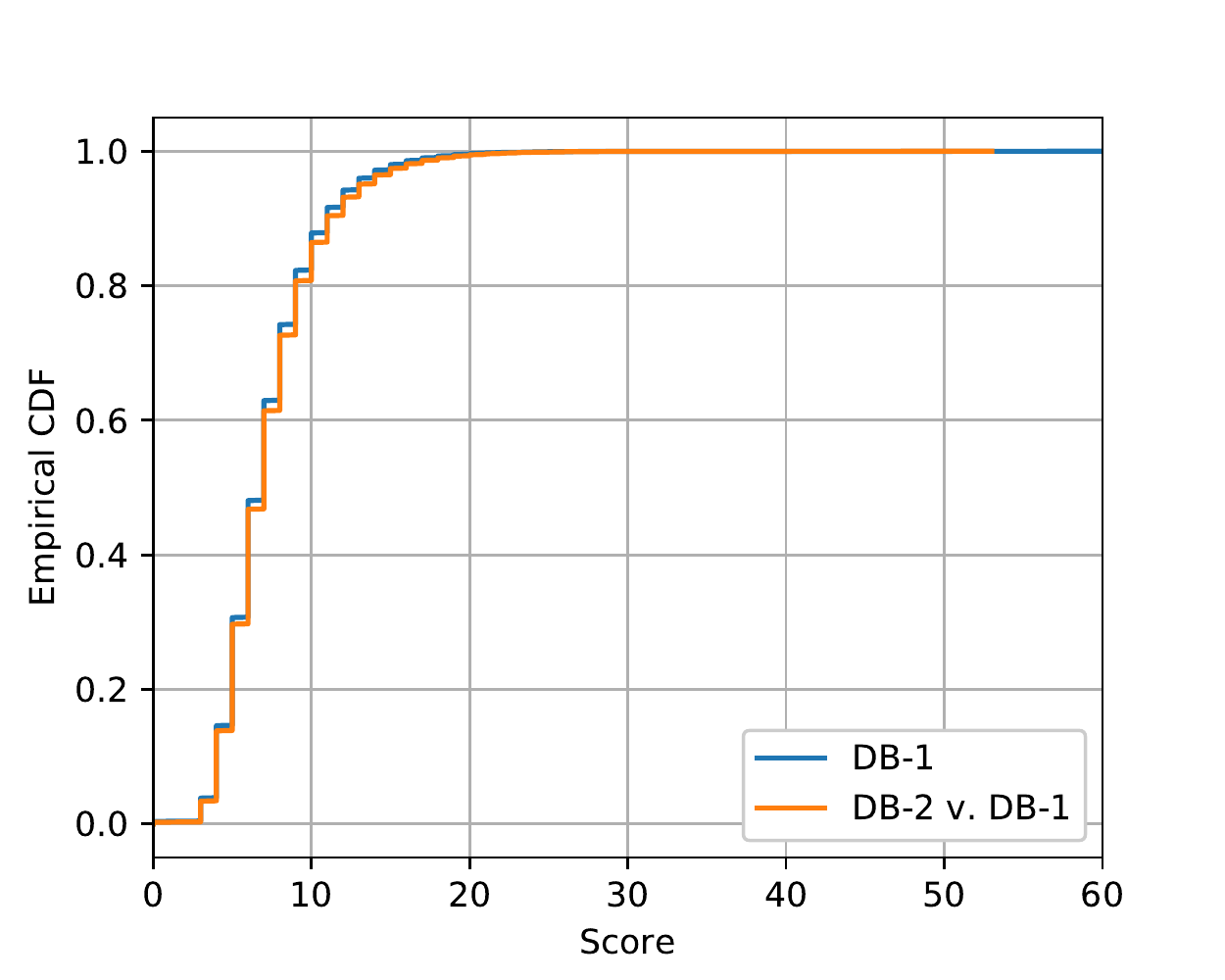}}
  \centerline{(d) CDFs for (a) and (b)}\medskip
\end{minipage}
\caption{Imposter score distributions for: (a) bonafide (DB-1), (B) synthetic (DB-2) versus bonafide (DB-1), (c) synthetic only (DB-2), and (d) empirical CDFs for (a) and (b).}
\label{fig:Imposters}
\end{figure}

To construct the empirical, genuine, and imposter distributions for the bonafide samples, we randomly select 1000 non-mated pairs for each fingerprint, resulting in 72 million non-mated scores, and we utilized all of the 3.8 million mated pairs. Figure \ref{fig:Imposters} (a) depicts the imposter distribution of the bonafide fingerprints (DB-1). We select a conservative False Accept Rate (FAR) of $0.001\%$ to evaluate the uniqueness of the synthetic fingerprints. Within the bonafide fingerprints (DB-1), this FPR translates into the True Positive Rate (TPR) of 0.873 at a score threshold of 41. Note that this threshold is in line with the rule of thumb threshold (40) suggested for the BOZORTH3 matcher \cite{ko_users_2007}. We use the same threshold (41) to evaluate the uniqueness of the synthetically generated fingerprints (DB-2). The synthetically generated fingerprints in DB-2 are compared with every bonafide fingerprint used for training the CFG (DB-1). This process produces more than 3 billion match scores. We randomly select 40 million of those to construct an empirical imposter match score distribution between synthetic fingerprints (DB-2) and bonafide fingerprints (DB-1). Figure \ref{fig:Imposters} (b) illustrates this imposter distribution. 

Interestingly the empirical distribution of synthetic fingerprints do not have the long tail that we observed in the imposter score distribution of the bonafide fingerprints (DB-1). Enforcing the same matching threshold (41) resulted in 190 false matches (maximum score of 53) as opposed to 330 false matches (maximum score of 286) observed in DB-1. Since the imposter distributions are not normally distributed, we conduct a non-parametric independent samples Kolmogorov-Smirnov test between the empirical imposter distributions to confirm the distinctness of the synthetic fingerprints.  We perform the test between the empirical imposter distribution observed by matching bonafide fingerprints (DB-1) and the empirical imposter distribution observed by matching synthetic to bonafide fingerprints (DB-2 v. DB-1) with an alternative hypothesis that the imposter distribution of DB-2-v-DB-1 is \emph{less} than DB-1. The Kolmogorov-Smirnov test $(D(40M) = 0.0158, p {>} 0)$ indicates rejecting the null hypothesis in favor of the alternative hypotheses. As a result, we can confirm that the synthetic fingerprints generated by the CFG (DB-2) are distinct from bonafide samples in the training dataset (DB-1) and do not reveal the identities presented in the training dataset (DB-1).

Finally, we also compared every synthetic fingerprint to every other synthetic fingerprint in DB-2 and randomly selected 40M comparisons. Figure \ref{fig:Imposters} (C) shows the empirical match score distribution for DB-2. We can observe that the distribution of match scores for synthetic fingerprints from DB-2 resembles that of bonafide in DB-1. However, the imposter distribution of DB-2 has a much longer tail that reveals an overlap of identities in the generated fingerprints. Enforcing the same threshold as DB-1 (41) results in 302,285 false matches out of 40M comparisons, compared to 330 in DB-1. Clearly there are multiple samples from the same identity in DB-2. This is expected as the CFG is trained in an unsupervised manner (without identity labels) and we did not force the generator to generate fingerprints by varying the disentangled identity portion of the learned latent variable ($W$). However, the tests also show that out of 50K samples 11.5K samples have no comparisons with match score above 41 and can be considered as synthetically generated unique identities. In our future work, we aim to train the next generation of the model using identity, finger, and quality labels. This would allow researchers to have granular control over those factors in the generated fingerprints.

\subsection{Quality, Diversity, and Fingerprint Metrics}
\label{sec:Metrics}

\begin{figure}[ht]
\centering
\centerline{\includegraphics[width=9.3cm]{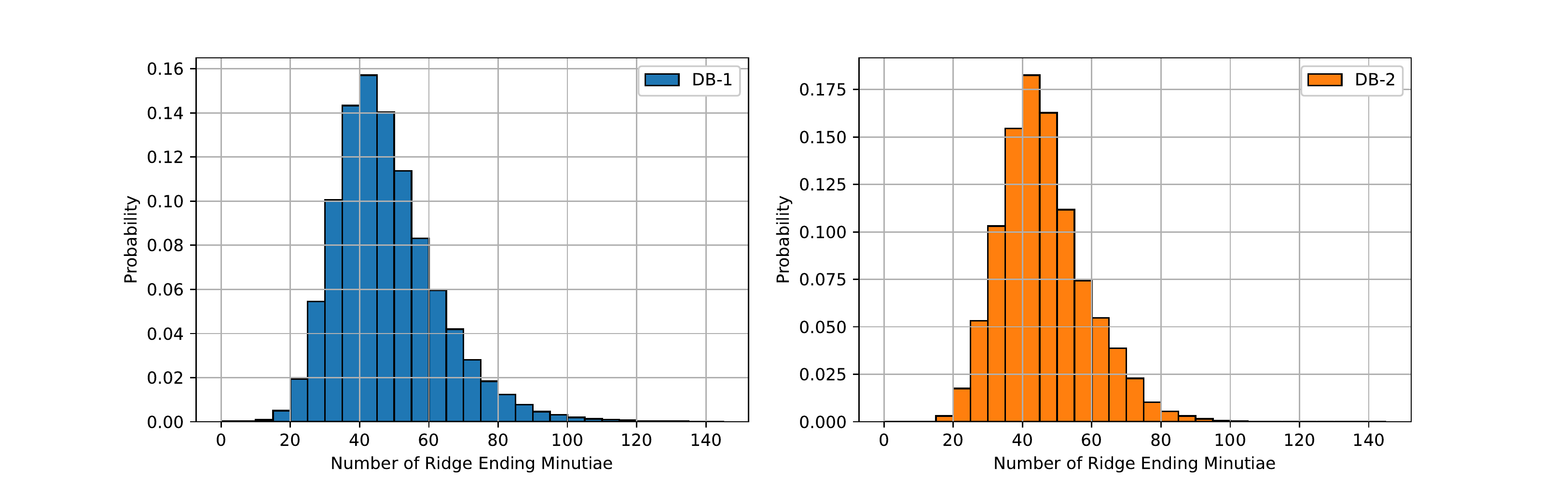}}
\caption{Number of ridge ending minutiae for DB-1 (Left) and DB-2 (Right)}
\label{fig:RIG}
\end{figure}

Table \ref{table:Finger_Metrics} presents fingerprint metrics for the synthetic and bonafide fingerprints. We observe high standard deviation in the evaluated metrics of the synthetic fingerprints. As an example, Figure \ref{fig:RIG} illustrates the distribution of Ridge Ending Minutiae in the training (DB-1) and synthetic (DB-2) fingerprints. This suggests that the CFG is generating fingerprints with diverse minutiae configurations and is not suffering from the mode collapse issues associated with other IWGAN models \cite{fahim_lightweight_2020,minaee_finger-gan_2018}. Additionally, we observe that the mean and standard deviation of the fingerprint metrics from the synthetically generated fingerprints are close to that of the training dataset (DB-1). Consequently, we can confirm the diversity and quality of the synthetically generated fingerprints. However, our results indicate differences in the skewness and kurtosis between synthetic and bonafide fingerprint metrics. For future work, we believe integrating such metrics into the GAN-based fingerprint synthesis model could improve the model's performance even further.

Additionally, we evaluated our synthetically generated fingerprint (DB-2) using the trained PAD model. The PAD model classified $95.2\%$ of the synthetic fingerprint as bonafide fingerprints. The PAD test reaffirms the quality and fidelity of the synthetically generated fingerprints.
\section{Conclusions}
\label{sec:conclusion}
In this work, we developed a progressive growth-based fingerprint synthesis model capable of generating 
$512\times512$ pixels, synthetic, plain, impression fingerprints which are diverse, unique, and high-fidelity. Our results show that progressive growth-based GAN models do not suffer from the shortcomings associated with the previously proposed IWGAN-based fingerprint synthesis models. Our results confirm that the synthetic fingerprints generated by the CFG closely resemble the fingerprints in the training dataset in terms of fingerprint minutiae configurations, ridge-valley structure, and quality while not revealing the identities presented in the training dataset. Finally, we make both the CFG and a dataset of synthetically generated samples publicly available to allow other researchers to continue this work.

\section{Future Work}
\label{sec:Future}
In future work, we aim to improve the CFG by integrating quality metrics into the loss function. Additionally, we aim to provide our model with identity, finger, and quality labels. This would allow us to control such factors during the synthesis process. Finally, we aim to further investigate the usefulness of synthetic fingerprints in training and evaluating PAD models.

\bibliographystyle{IEEEbib}
\bibliography{finger.bib}
\end{document}